# *RNNAccel:* A Fusion Recurrent Neural Network Accelerator for Edge Intelligence

*Chao-Yang Kao[1,2], Huang-Chih Kuo[1,2], Jian-Wen Chen[1,2], Chiung-Liang Lin[1,2], Pin-Han Chen[2], Youn-Long Lin[1]*

[1]Department of Computer Science, National Tsing Hua University
[2]Neuchips Corporation
{chaoyang_kao, huangchih_kuo, jianwen_chen, chiungliang_lin, kinny_chen}@neuchips.ai, ylin@cs.nthu.edu.tw

## ABSTRACT

Many edge devices employ Recurrent Neural Networks (RNN) to enhance their product intelligence. However, the increasing computation complexity poses challenges for performance, energy efficiency and product development time. In this paper, we present an RNN deep learning accelerator, called *RNNAccel*, which supports Long Short-Term Memory (LSTM) network, Gated Recurrent Unit (GRU) network, and Fully Connected Layer (FC)/ Multiple-Perceptron Layer (MLP) networks. This RNN accelerator addresses (1) computing unit utilization bottleneck caused by RNN data dependency, (2) inflexible design for specific applications, (3) energy consumption dominated by memory access, (4) accuracy loss due to coefficient compression, and (5) unpredictable performance resulting from processor-accelerator integration. Our proposed RNN accelerator consists of a configurable 32-MAC array and a coefficient decompression engine. The MAC array can be scaled-up to meet throughput requirement and power budget. Its sophisticated off-line compression and simple hardware-friendly on-line decompression, called NeuCompression, reduces memory footprint up to 16x and decreases memory access power. Furthermore, for easy SOC integration, we developed a tool set for bit-accurate simulation and integration result validation. Evaluated using a keyword spotting application, the 32-MAC RNN accelerator achieves 90% MAC utilization, 1.27 TOPs/W at 40nm process, 8x compression ratio, and 90% inference accuracy.

## 1. INTRODUCTION

With the advance of deep learning neural network computing, more researches are focusing on how to deploy trained neural network models to application systems. However, bulky models would rapidly increase computation complexity for edge devices. RNN, LSTM [10], and GRU [11] are popular Artificial Neural Network for sequence-data processing. Several LSTM and GRU accelerators have been realized in FPGA or ASIC. For FPGA-based design, Han [1] proposed a model pruning and quantization technique to reduce the number of MAC operations of LSTM. Zhang [6] proposed a pipelined architecture together with a multiple-tile matrix-vector multiplication engine to improve the computing efficiency. Gao [2] developed a delta network algorithm to increase the sparsity of activation results and then realized a high-throughput GRU accelerator in FPGA.

Several ASIC implementations of LSTM accelerators have been proposed for high performance and low power consumption. Shin [3] replaced matrix-vector product with quantization tables at the expense of prediction accuracy, resulting in reduced external memory bandwidth. Conti [4] kept weight parameters in multiple SRAM banks and tiled multiple engines together to support large LSTM models, which need high computation power. Park [5] proposed a new sparse matrix format to minimize the load imbalance for fast sparse matrix-vector operation. Yazdani [7] designed a dynamic reconfigurable matrix-vector engine to improve the hardware utilization and proposed a scheduling scheme to reduce computational dependency of LSTM.

To deploy RNN style neural network models in the edge, we have to address several issues. First, the accelerator should support various types of networks such as LSTM, GRU, FC/MLP, and arbitrary composition of above. Second, it should support multiple concurrent models for an application. Third, it should support model weight coefficients off-line compression and on-chip decompression for cost sensitive applications.

The rest of the paper is organized as following. In Section 2, we present our proposed design including the accelerator architecture, the decompression circuit, and the companion software. In Section 3, we evaluate the proposed design and compare it with some previous results. Finally, we draw conclusion and point to possible directions for future research in Section 4.

## 2. PROPOSED DESIGN

We first describe the proposed accelerator architecture in Section 2.1. We then propose a compression algorithm in Section 2.2. In Section 2.3, we present our multi-function activation circuit. In Section 2.4, we describe our companion software development kits.

### 2.1. Proposed Architecture

Figure 1 depicts a top-level block diagram of the proposed *RNNAccel* accelerator. It consists of five functional blocks and a local memory pool.

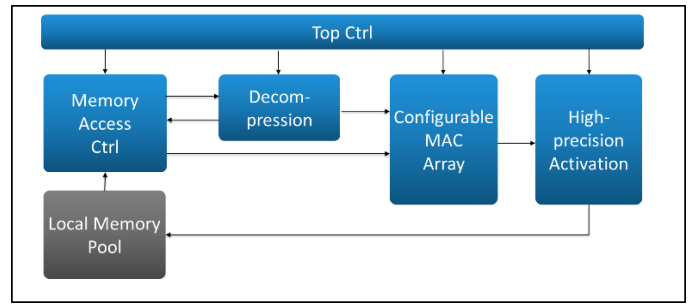

Figure 1 Top-level block diagram of *RNNAccel*

The Top Ctrl unit is responsible for interfacing the accelerator and the processor. It issues control signals such as memory access enable, network type, and activation type. The Memory Access Ctrl unit reads inputs and weights, and then writes the inference results. To simplify SOC integration, we make it compatible with widely used I/O interface such as MMIO and AMBA [13].

We support both compressed and uncompressed weight coefficients. If compression is disabled, the Memory Access Ctrl unit directly sends weights to the MAC Array unit. On the other hand, if compression is enabled, the Decompression unit receives the compressed bit-streams and decompresses them into weights. Section

2.2 describes the proposed compression/decompression algorithm in more detail.

The MAC Array unit performs matrix-vector multiplications. Each MAC in the array performs 16-bit x 8-bit multiply-and-accumulate operation. To support applications that use 16-bit weights to achieve high precision, we can combine two 16-bit x 8-bit MACs into a 16-bit x 16-bit MAC. Moreover, the number of MACs in the MAC array unit is also scalable.

The Activation unit takes the MAC results as input and produces activation values as outputs. Because several activation functions (i.e., Sigmoid and Tanh) involves exponential function, to minimize approximate error is important when designing an approximation circuit. The proposed high-precision Activation unit will be further described in Section 2.3.

The Local Memory Pool contains several memory banks for storing intermediate data, such as cell values of an LSTM. The number of memory banks is also configurable to meet various application requirements.

## 2.2. Weight Coefficient Compression

Energy efficiency is an important metrics for edge devices. Previous study [14] shows that memory access usually dominates edge AI inference power consumption. Many neural network compression approaches, such as pruning and quantization have been proposed to reduce memory footprint. Pruning is proven effective for large neural networks (millions of parameters) but less effective for small and compact (kilos of parameters) neural networks. Moreover, no compression method is designed specific for RNN. Therefore, we propose a novel network compression called *NeuCompression* to deal with compact RNN for edge applications.

*NeuCompression* is composed of a sophisticated off-line compression algorithm and a simple on-line decompression circuit. It adopts a user-chosen fixed compression ratio to simplify the decompression process and thus minimize the hardware cost. *NeuCompression* currently supports three fixed compression rate (5.3x, 8x and 16x) to make a trade-off between quality loss and memory footprint. We can easily add more compression rate in the future.

## 2.3. Activation Function Approximation Circuit

We propose a multi-mode pipelined approximate circuit which can support Tanh, Sigmoid, Softsign, and Relu activation functions. It uses a piecewise linear function to approximate Tanh, Sigmoid and Softsign function. We reduce the number of pieces by exploiting the symmetric characteristics of Tanh and Softsign. In addition, Tanh and Sigmoid can share the approximate functions through formula transformation. Simulation results show that the maximum error of our approximation function is below 0.0002.

## 2.4. Software Development Tools

Figure 2 depicts a set of software tools to facilitate SOC integration of *RNNAccel*. *NN Compiler* analyzes neural networks trained in Pytorch or Tensorflow frameworks and extracts information of neural network architecture parameters and weight coefficients.

*Behavior Simulator* provides bit-level accurate evaluation of *RNNAccel* performance evaluation (cycle-level throughput, fixed-point computation accuracy, and compression ratio and quality loss tradeoff) at the initial design stage.

*NeuCompression* compresses neural network parameters according to user specified compression rate and then generates encoded parameter bit-stream. It also provides decoded parameters to *Behavior Simulator* for compression quality evaluation. *Validation Tool* compares inference results of *Behavior Simulator* and *RNNAccel* to verify the correctness of SoC integration.

## 3. RESULTS AND COMPARISON

Synthesized to run at 250MHz using a TSMC 40nm CMOS cell library, the proposed accelerator uses 184.1K gates of logic and 12K bytes of local SRAM memory. To evaluate its performance, we run a GRU-style network proposed in the Hello Edge Keyword Spotting [15] that performs keyword spotting using Google Speech Command Dataset [9]. The network consists of one GRU layer and one FC layer, as depicted in Figure 3. The network takes 10 pre-processed voice data and outputs 12 results, which indicates the possibility of 10 keywords, unknown word, and silence, respectively.

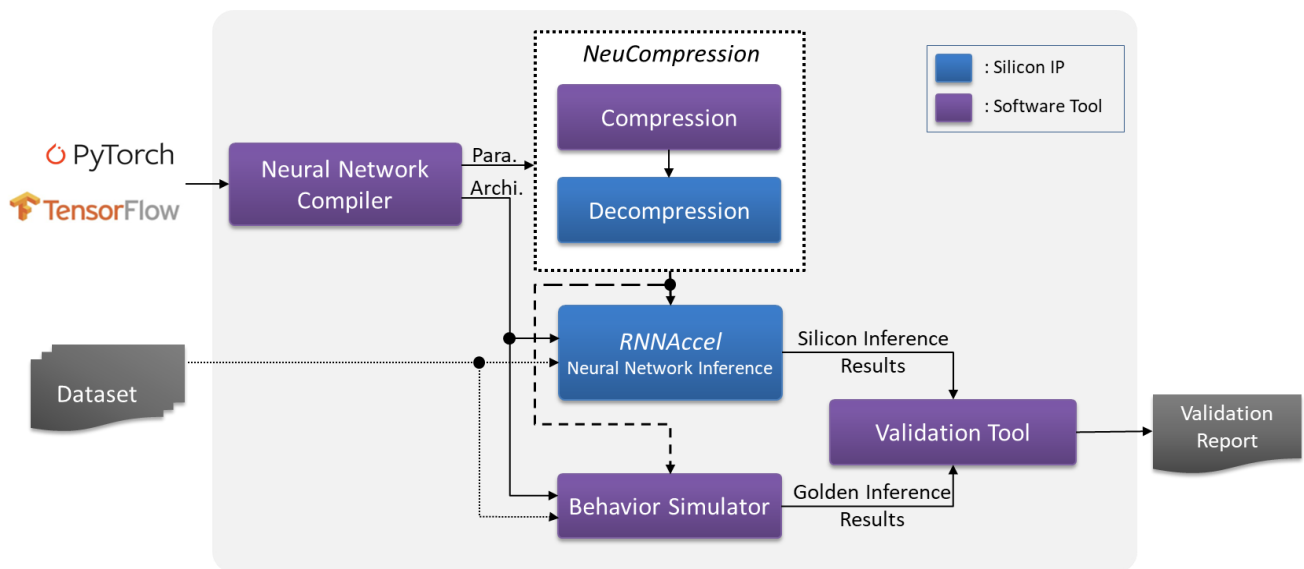

Figure 2 *RNNAccel* 's Software Development Tools

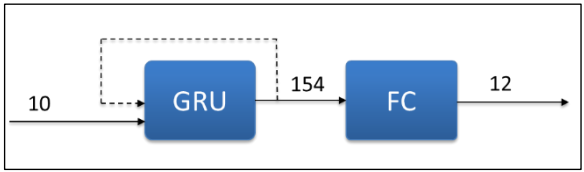

Figure 3 A Keyword Spotting GRU

Experiment result shows that the proposed accelerator delivers up to 91K inferences per second and consumes 12.6mW (4mW in the Local Memory Pool unit and 8.6mW in the functional units) when running at 250MHz. Figure 3 compares our accelerator with several previous works. The proposed accelerator supports most widely used recurrent networks and delivers up to 1.27 TOPS/W energy efficiency with 32 MACs. The proposed accelerator uses few MACs to deliver high energy efficiency because the high hardware utilization. For example, the utilization of the MAC Array unit is 90% for the GRU network shown in Figure 3.

To evaluate the effectiveness of NeuCompression, we take two RNN applications as benchmarks to compare pruning, quantization, and compression under various compression ratio target. The GRU-style neural network, which is shown in Figure 3. with 78K parameters to detect keywords. The Atrial Fibrillation Detection [16] system adopts a Bi-LSTM-style neural network with 40K parameters to detect abnormality in ECG signals. Experiment results in Figure 4 and Figure 5 show *NeuCompression* delivers outstanding performance in these two RNN-based edge applications. In these small compact networks, both pruning and quantization result in substantial accuracy loss. Even retraining cannot recover much of the loss. On the contrary, our *NeuCompression* technology keeps the accuracy loss negligible at 8x compression (i.e., 4 bits per weight). For more aggressive 16x compression (i.e., 2 bits per weight), the accuracy is still acceptable after retraining.

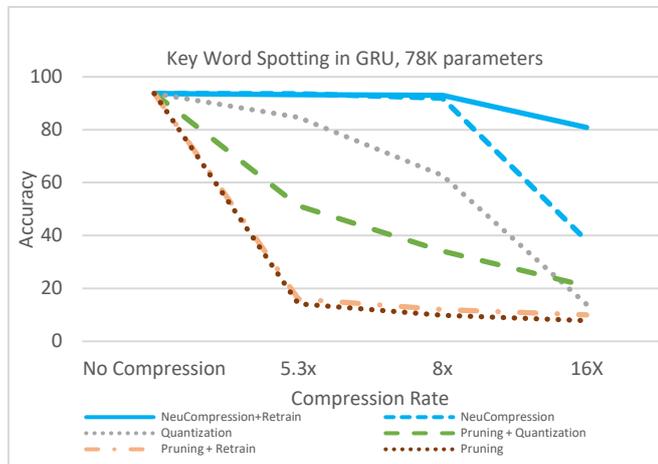

Figure 4 NeuCompression Maintains Higher Accuracy after 8x Compression for a 78K-parameter GRU Network

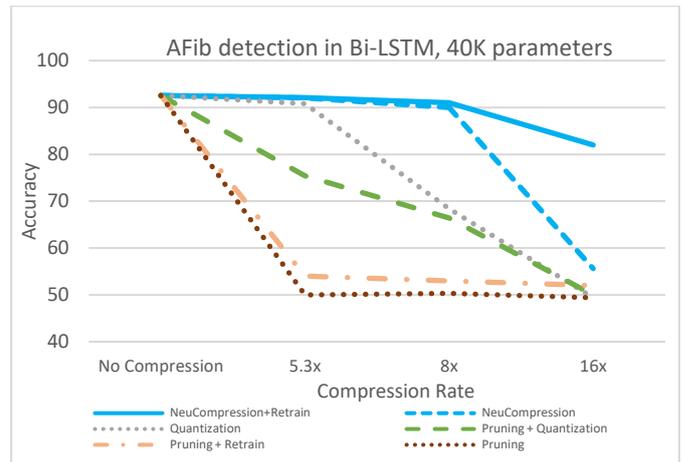

Figure 5 NeuCompression Maintains Higher Accuracy after 8x Compression Ratio for AFib Detection Application in Bi-LSTM

## 4. CONCLUSION AND FUTURE WORK

We have proposed an RNN silicon IP that includes a LSTM/GRU/FC inference accelerator with *NeuCompression* and an SDK. It demonstrates advantages in performance, silicon area, energy efficiency, and ease of integration and verification. Synthesized into a TSMC 40nm CMOS cell library, the IP consumes 184K gates to run at 250MHz. By *NeuCompression* offline compression and on-the-fly decompression, memory footprint can be reduced by 8x without suffering from quality loss. Simulated using a keyword spotting application with a fusion neural network combines one GRU layer and one FC layer, the *RNNAccel* can deliver (1) $\geq$ 90% computing unit utilization, (2) flexible for multiple different applications with configurable MAC array, (3) high energy efficiency by reducing memory footprint (1.27 GOPS/W with 32MACs), (4) $\geq$ 90% inference accuracy with 8x compression ratio, (5) predictable performance by Behavior Simulator software. The proposed solution can be easily integrated into popular SOC systems to speed up neural network computation.

To make the IP useful for more applications, we would like to optimize the development flow. Figure 6 shows the development flow from trained model to application development. Once a trained neural network model is converted to the Open Neural Network Exchange (ONNX) [13] format, the ONNC compiler produces a loadable file according to our instruction set architecture (ISA). An application then integrates this loadable file to access the accelerator through its drivers. With this standard open neural network development flow, edge device developers can do their jobs in a modular fashion.

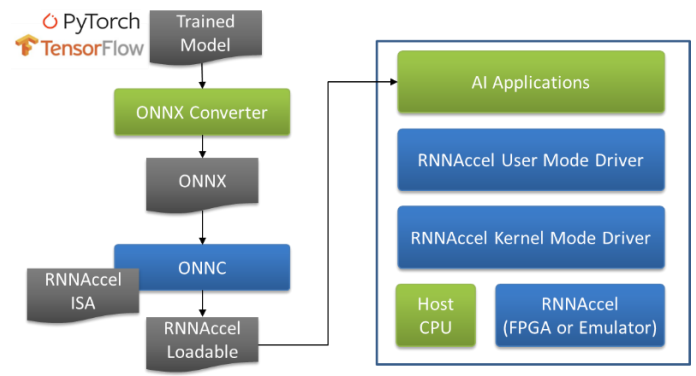

Figure 6 *RNNAccel* application development flow with Open Neural Network platform

**Table 1 Comparison result**

| Works | Realized Recurrent Model | Implementation | Frequency (MHz) | Number of PEs | Power (W) | Throughput (GOPS) | Efficiency (GOPS/W) |
|---|---|---|---|---|---|---|---|
| **Han [1]** | LSTM | FPGA | 200 | 1,024 | 41 | 282 | 7 |
| **Gao [2]** | GRU | FPGA | N/A | 768 | 7.3 | 1198 | 164 |
| **Shin [3]** | LSTM | ASIC@65 nm | 200 | 64 | 0.021 | 25.6 | 1,219 |
| **Yazdani [6]** | LSTM | ASIC@90 nm | 500 | 64,000 | 39.1 | 29800 | 762 |
| **Proposed** | LSTM/GRU/FC | ASIC@40 nm | 250 | 32 | 0.0126 | 16 | 1,270 |